%% file: main.tex
\documentclass[10pt,twocolumn,letterpaper]{article}

\usepackage[pagenumbers]{cvpr} 


\definecolor{cvprblue}{rgb}{0.21,0.49,0.74}
\usepackage[pagebackref,breaklinks,colorlinks,allcolors=cvprblue]{hyperref}

\usepackage{url}            
\usepackage{microtype}      
\usepackage{xcolor}         
\usepackage{colortbl}
\usepackage{multirow}
\usepackage{amsmath}
\usepackage{amsfonts}
\usepackage{dsfont}
\usepackage{graphicx}
\usepackage{pifont} 
\usepackage{fontawesome} 

\newcommand{\cellbest}{\cellcolor{red!40}}
\newcommand{\cellsecond}{\cellcolor{orange!40}}
\newcommand{\cellthird}{\cellcolor{yellow!40}}

\title{RelayGS: Reconstructing Dynamic Scenes with Large-Scale and Complex Motions via Relay Gaussians}

\author{
  Qiankun~Gao\textsuperscript{1,~2},~~ Yanmin~Wu\textsuperscript{1},~~ Chengxiang~Wen\textsuperscript{1},~~ Jiarui~Meng\textsuperscript{1},~~ Luyang~Tang\textsuperscript{1,~2,~3},~~\\  Jie~Chen\textsuperscript{1,~2~\href{mailto:chenj@pcl.ac.cn}{\faEnvelopeO}},~~ Ronggang~Wang\textsuperscript{1,~2,~3},~~ \href{https://jianzhang.tech}{Jian~Zhang}\textsuperscript{1,~2,~3~\href{mailto:zhangjian.sz@pku.edu.cn}{\faEnvelopeO}}\\
  \normalsize{\textsuperscript{1}School of Electronic and Computer Engineering, Peking University~~~~\textsuperscript{2}Peng Cheng Laboratory} \\
  \normalsize{\textsuperscript{3}Guangdong Provincial Key Laboratory of Ultra High Definition Immersive Media Technology}, \\
  \normalsize{Peking University Shenzhen Graduate School}\\
}

\begin{document}
\maketitle
\input{sec/0_abstract}    
\input{sec/1_introduction}
\input{sec/2_related_work}
\input{sec/3_preliminary}
\input{sec/4_method}
\input{sec/5_experiment}

\input{sec/6_conclusion}
{
    \small
    \bibliographystyle{ieeenat_fullname}
    \bibliography{main}
}

\input{sec/X_suppl}

\end{document}

%% file: sec/0_abstract.tex
\begin{abstract}
Reconstructing dynamic scenes with large-scale and complex motions remains a significant challenge. Recent techniques like Neural Radiance Fields and 3D Gaussian Splatting (3DGS) have shown promise but still struggle with scenes involving substantial movement. This paper proposes \textbf{RelayGS}, a novel method based on 3DGS, specifically designed to represent and reconstruct highly dynamic scenes. Our RelayGS learns a complete 4D representation with canonical 3D Gaussians and a compact motion field, consisting of three stages. First, we learn a fundamental 3DGS from all frames, ignoring temporal scene variations, and use a learnable mask to separate the highly dynamic foreground from the minimally moving background. Second, we replicate multiple copies of the decoupled foreground Gaussians from the first stage, each corresponding to a temporal segment, and optimize them using pseudo-views constructed from multiple frames within each segment. These Gaussians, termed \textbf{Relay Gaussians}, act as explicit relay nodes, simplifying and breaking down large-scale motion trajectories into smaller, manageable segments. Finally, we jointly learn the scene’s temporal motion and refine the canonical Gaussians learned from the first two stages. We conduct thorough experiments on two dynamic scene datasets featuring large and complex motions, where our RelayGS outperforms state-of-the-arts by more than 1 dB in PSNR, and successfully reconstructs real-world basketball game scenes in a much more complete and coherent manner, whereas previous methods usually struggle to capture the complex motion of players.
\end{abstract}

%% file: sec/1_introduction.tex
\section{Introduction}
Dynamic scene reconstruction plays a pivotal role in a wide range of applications requiring immersive and interactive environments, such as virtual reality, metaverse, and free-viewpoint videos. However, achieving high-fidelity reconstruction of dynamic scenes with large-scale and complex motions from multi-view videos remains highly challenging. 

The recently emerged Gaussian Splatting (3DGS)\cite{3dgs} has advanced 3D reconstruction, enhancing efficiency and quality compared to its predecessor, Neural Radiance Fields (NeRF)\cite{nerf}. Using Gaussian ellipsoids as explicit 3D primitives, 3DGS achieves real-time $1080p$ rendering via a rasterized pipeline. Dynamic extensions of 3DGS~\cite{4d-gaussian-splatting,deformabl3dgs,modgs,sc-gs,gags,splatfields,sp-gs,dynmf,gaussian-flow,mags} typically combine canonical representations with implicit motion fields, similar to dynamic NeRFs~\cite{d-nerf,nerfies,hypernerf}. While effective for small-scale motions, these methods face challenges with large, complex motions in real-world scenarios, such as sports events with fast-moving players. The primary limitation stems from the coupling of canonical Gaussian learning with neural motion fields, which complicates optimization. Neural networks not only find it challenging to predict large motions but also tend to overfit the dominant small motions in the scene, limiting their ability to model extensive complex movements.

A key approach to addressing large-scale and complex motion is to decouple the highly dynamic foreground from the minimally moving background. By isolating the foreground, we can better capture complex motion trajectories, while minimizing background interference. MLPs efficiently represent the background’s motion dynamics; however, the main challenge lies in modeling large, non-rigid foreground motions, which we tackle by decomposing these trajectories into shorter, simpler segments.

In this paper, we propose \textbf{RelayGS} to reconstruct dynamic scenes with large-scale, complex motions from multi-view videos. Our goal is to achieve a complete 4D representation, comprising a set of explicit canonical 3D Gaussians and a compact motion field. The core idea is to simplify complex motion trajectories during the learning of canonical 3D Gaussians, laying the foundation for the subsequent joint learning of the 3D Gaussians and the motion field. Specifically, our method unfolds in three progressive stages:

$\bullet$ \textbf{I)} 
We begin by learning a static initial 3DGS from all frames without considering temporal changes, primarily capturing the shared background. To also represent the foreground, we introduce a \textit{learnable mask} to distinguish high-dynamic foreground Gaussians from low-dynamic background Gaussians. All Gaussians are used to render the first frame, while only those with mask = 1 are used in subsequent frames, yielding a coarse representation of both the background and initial foreground while effectively decoupling the two (see Sec.~\ref{sec:method:stage1} for details).

$\bullet$ \textbf{II)} 
Ideally, each foreground Gaussian would follow a complex motion trajectory over time, but achieving this directly is challenging. Instead, we replicate multiple copies of the decoupled foreground Gaussians from the first stage, each corresponding to a temporal segment. To further optimize and densify these copies, we construct pseudo-views using selected frames from the corresponding segment. These foreground Gaussians, which we term \textbf{Relay Gaussians}, serve as explicit, discrete nodes along the idealized motion trajectory, effectively simplifying and approximating the complex, large-scale trajectory by breaking it down into smaller, manageable segments. (see Sec.~\ref{sec:method:stage2} for details). \textit{This can be regarded as temporal densification, analogous to the spatial densification in 3DGS} (refer to Sec.~\ref{sec:supp1} in \textit{supp.}). 

$\bullet$ \textbf{III)} 
Finally, we jointly optimize the canonical Gaussians learned in the previous stages together with a compact motion field to achieve a complete 4D representation. Though our method is not limited to a specific motion field model, we follow 4D-GS~\cite{4dgs} in this paper by adopting HexPlane~\cite{hexplane} and lightweight MLPs, with several key modifications to better capture large, complex motions. Specifically, for the foreground \textbf{Relay Gaussians}, we employ an additional set of MLPs and introduce a learnable scaling factor for position offsets as they may require a larger range that cannot be fully captured by the MLP’s predictions alone. These improvements allow the foreground Relay Gaussians to more accurately represent the dynamic and complex motions in the scene. (see Sec.~\ref{sec:method:stage3} for details).

We conducted thorough experiments to validate the effectiveness of our \textbf{RelayGS}. 
On the PanopticSports dataset~\cite{panoptic_studio}, featuring large-scale motions, our method achieves a \textbf{1 dB} PSNR improvement over previous state-of-the-arts. On the more challenging VRU Basketball Games dataset~\cite{vru}, our method reconstructs scenes with greater completeness and coherence, where prior methods struggle to capture the dynamic foreground content with complex motions. The contributions of this paper are summarized as follows:

\noindent\ding{113} 
We introduce a simple learnable mask that effectively decouples high dynamic foreground and low dynamic background Gaussians without relying on additional priors, while learning a more accurate and complete fundamental 3DGS representation of the dynamic scene.

\noindent\ding{113} We propose the temporal Relay Gaussians to decompose large-scale and complex motion trajectories into smaller, more manageable motion segments, simplifying the representation and learning of complex dynamics. 

\noindent\ding{113} We utilize distinct MLPs to predict motion changes for background Gaussians and foreground Relay Gaussians, along with a learnable scaling factor for the position changes of Relay Gaussians, enabling accurate capture of larger and more complex motions.

\noindent\ding{113} We conduct experiments on real-world dynamic scene datasets featuring large-scale, complex motions, where our RelayGS significantly outperforms previous state-of-the-art methods, achieving a \textbf{1 dB} improvement in PSNR on PanopticSports dataset and delivering more complete and coherent reconstructions of complex, large-scale foreground motions.

%% file: sec/2_related_work.tex
\section{Related Work}

\noindent\textbf{Dynamic Scene Modeling.}
Early methods~\cite{d-nerf, nerfies, hypernerf, n3dv,tensor4d,hexplane,kplanes} based on Neural Radiance Fields (NeRF) model deformation fields to map canonical spaces to dynamic frames but suffer from high computational costs due to dense sampling.
Recent approaches have shifted toward the more efficient 3D Gaussian Splatting (3DGS). A straightforward strategy expands Gaussian primitives to 4D, as seen in 4DGS~\cite{4d-gaussian-splatting} and Rotor-4DGS~\cite{rotor4dgs}. Other methods decouple the dynamic scene into a canonical 3DGS and a temporal motion field. Deformable3DGS~\cite{deformabl3dgs} uses deep MLPs to predict Gaussian motion, while 4D-GS~\cite{4dgs} enhances this framework with multi-resolution HexPlane and lightweight MLPs for improved efficiency. SC-GS~\cite{sc-gs} assumes motion is driven by key points, predicting time-varying transformations through a deformation MLP and interpolating them to generate a coherent motion field.
Additionally, online frame-by-frame learning methods incrementally model dynamic changes. Dynamic3DGS~\cite{dynamic3dgs} updates Gaussian positions and rotations at each timestamp, while 3DGStream~\cite{3dgstream} employs a NGP~\cite{ngp} to manage transformations efficiently. HiCoM~\cite{hicom} leverages the non-uniform distribution and local consistency to enable fast and accurate motion learning across frames.

\noindent\textbf{Dynamic-Static Decoupling.} S4D~\cite{s4d}, EgoGaussian~\cite{egogaussian}, and SC-4DGS~\cite{sc-4dgs} use pre-trained segmentation models to generate 2D motion masks for separating dynamic and static content. Compact-D3DGS~\cite{compact-d3dgs} and GauFRe~\cite{gaufre} rely on optical flow. In contrast, our method uses a learnable mask to decouple high-dynamic foreground from low-dynamic background, eliminating reliance on pre-trained motion priors and enhancing adaptability to complex motions.

%% file: sec/3_preliminary.tex
\section{Preliminaries}

\paragraph{3D Gaussian Splatting.}
3D Gaussian Splatting~\cite{3dgs} explicitly represents scenes using anisotropic 3D Gaussian primitives, mathematically formulated as:
\begin{equation}
    \mathcal{G}(\mathbf{x})=e^{-\frac{1}{2}(\mathbf{x} - \boldsymbol{\mu})^T\mathbf{\Sigma}^{-1}(\mathbf{x} - \boldsymbol{\mu})}, \quad
    \mathbf{\Sigma} = \mathbf{R}\mathbf{S}\mathbf{S}^T\mathbf{R}^T,
\end{equation}
where the mean vector $\boldsymbol{\mu}$ and covariance matrix $\mathbf{\Sigma}$ respectively characterize the central position and geometric shape. The matrix $\mathbf{\Sigma}$ is decomposed into a scaling matrix $\mathbf{S} = \text{diag}(s_x, s_y, s_z)$ and a rotation matrix $\mathbf{R} \in SO(3)$, further simplified as a vector $\mathbf{s} \in \mathbb{R}^3$ and a quaternion $\mathbf{q} \in \mathbb{R}^4$, to ensure physical meaning and facilitate optimization. Each Gaussian is associated with an opacity $o$ and spherical harmonics $\mathbf{h}$ representing color.

Rendering is performed by blending the contributions of $N$ overlapping Gaussian primitives at each pixel, taking into account their depth-ordering to ensure correct compositing, expressed as: 
\begin{equation}
    C = \sum_{i \in N} \mathbf{c}_i \alpha_i \prod_{j=1}^{i-1} (1 - \alpha_j),
\end{equation}
where $\mathbf{c}_i$, $\alpha_i$ represents the color and blending weight of the $i^{th}$ Gaussian, respectively.
The training alternates between parameter optimization and density control. Parameter optimization is supervised by the $\mathcal{L}_1$ loss and D-SSIM term:
\begin{equation}
    \mathcal{L}=(1-\lambda)\mathcal{L}_1+\lambda\mathcal{L}_\text{D-SSIM}.
\end{equation}

\paragraph{4D Gaussian Splatting.}
4D-GS~\cite{4dgs} extends 3D Gaussian Splatting (3DGS) by incorporating a deformation field to model dynamic scenes. The deformation field is implemented through a HexPlane~\cite{hexplane} encoding module $\mathcal{H}$ and a set of lightweight MLPs. Based on the Gaussian center position $\boldsymbol{\mu} = (x,y,z)$ and a given time $t$, $\mathcal{H}$ outputs a feature encoding $\mathbf{f}$, which is then fed into separate MLPs to predict changes in Gaussian attributes such as position $\boldsymbol{\mu}$, scale $\mathbf{s}$, rotation $\mathbf{r}$, and opacity $o$, represented as follows:
\begin{equation}
\begin{aligned}
    \Delta \boldsymbol{\mu} &= \phi_{\boldsymbol{\mu}}(\mathbf{f}), \quad & \Delta \mathbf{r} &= \phi_\mathbf{r}(\mathbf{f}), \\
    \Delta \mathbf{s} &= \phi_\mathbf{s}(\mathbf{f}), \quad & \Delta \mathbf{o} &= \phi_\mathbf{o}(\mathbf{f}).
\end{aligned}
\end{equation}
The deformed 3D Gaussian is expressed as:
\begin{equation}
    \mathcal{G}^\prime=\{
        \boldsymbol{\mu} + \Delta \boldsymbol{\mu}, 
        \mathbf{s} + \Delta \mathbf{s}, 
        \mathbf{r} + \Delta \mathbf{r}, 
        o +\Delta o, 
        \mathbf{h}
    \}.
\end{equation} 
At time $t$, the 3D Gaussian $\mathcal{G}$ in the scene will be replaced by the deformed 3D Gaussian $\mathcal{G}^\prime$ for rendering.

The training process consists of two stages. The first stage serves as a warm-up, optimizing static scenes using only 3D Gaussians. In the second stage, the HexPlane, MLPs, and 3D Gaussians are jointly optimized. The loss function includes an $\mathcal{L}_1$ loss and a grid-based total variation loss $\mathcal{L}_{tv}$:
\begin{equation}
    \mathcal{L} = \mathcal{L}_1 + \mathcal{L}_{tv}.
    \label{eq:4dgs_loss}
\end{equation}

%% file: sec/4_method.tex
\section{Methodology}
The proposed method, \textbf{RelayGS}, is designed to effectively tackle the challenge of reconstructing dynamic scenes with large-scale and complex motions. The goal, similar to our baseline method 4D-GS~\cite{4dgs}, is to achieve a complete 4D representation consisting of a set of explicit canonical 3D Gaussians and a compact motion field. Our core idea is to simplify complex motion trajectories during the learning of canonical 3D Gaussians (detailed in Secs.~\ref{sec:method:stage1} and~\ref{sec:method:stage2}), thereby laying a strong foundation for the subsequent joint learning of canonical 3D Gaussians and the motion field (detailed in Sec.~\ref{sec:method:stage3}), as illustrated in Fig.\ref{fig:framework}. While our method is not limited to a specific motion field, we adopt HexPlane and lightweight MLPs, following 4D-GS, with several improvements to better accommodate complex motions. We introduce the three progressive stages in detail below.

\input{figure/pipeline}

\subsection{Stage 1: Initial Representation and Foreground-Background Decoupling}
\label{sec:method:stage1}

The primary goal of this first stage is to construct the fundamental 3D structure of the dynamic scene. Previous method~\cite{4dgs} initialize a set of static Gaussians from sparse point clouds and jointly optimize them using all given frames without considering temporal scene changes, \textit{i.e.}, treating it as a static scene for initialization. This approach effectively captures the relatively static background of the scene, but struggles with the highly dynamic foreground.

The highly dynamic foreground, due to its significant positional variations across frames, cannot be easily initialized. For instance, even if some Gaussians can model dynamic foreground objects in a specific frame, due to the large motion of the objects, they may cause inconsistencies in another frame, resulting in large rendering errors. Under this initialization paradigm, the Gaussians representing such foreground objects would be noisy or automatically pruned.

To address this limitation and learn the highly dynamic foreground simultaneously, we introduce a \textit{``learnable mask''} for each Gaussian primitive to indicate whether it belongs to the highly dynamic foreground or the relatively static background. The implementation of this mask follows the straight-through estimator~\cite{ste}, a technique widely adopted in previous works~\cite{compact3dgs,hac} to assess the importance of each Gaussian primitive for rendering quality in static scenes, enabling effective pruning and compression to reduce storage overhead. However, we are the first to leverage this approach in the context of dynamic scene reconstruction, using it to distinguish between foreground and background Gaussians. The formulation is expressed as:
\begin{equation}
    \begin{aligned}
        \mathbf{M}_n &= \text{sg}(\mathds{1}[\sigma(\mathbf{m}_n)>\epsilon]-\sigma(\mathbf{m}_n))+\sigma(\mathbf{m}_n) \\
                     &= \begin{cases}
            1, & \text{if } \sigma(\mathbf{m}_n) > \epsilon \\
            0, & \text{otherwise}
          \end{cases},
    \end{aligned}
\end{equation}
where $n$ is the index among all $N$ Gaussians, $\epsilon$ is the masking threshold, $\mathbf{m} \in \mathbb{R}^N$ is the learnable mask parameter, $\mathbf{M} \in \{0, 1\}^N$ is the generated binary masks,
$\text{sg}\left(\cdot\right)$ is the stop gradient operator, and $\mathds{1}\left[\cdot\right]$ 
and $\sigma\left(\cdot\right)$ are indicator and sigmoid function, respectively. It is important to note that, although $\mathbf{M}_n$ is binary, gradients can still be backpropagated to $\mathbf{m}_n$, allowing for optimization through gradient descent.

We use all Gaussians to render views for the first frame, ignoring the mask. However, when rendering other frames, we replace each Gaussian’s opacity with the following:
\begin{equation}
    \hat{\mathbf{o}}_n = \mathbf{M}_n \mathbf{o}_n,
    \label{eq:alpha_mask}
\end{equation}
where, $\mathbf{o}_n$ and $\hat{\mathbf{o}}_n$ are the opacity before and after applying the mask, respectively. The Gaussians representing highly dynamic foreground in the first frame incur a larger loss in other frames due to their movement, resulting in higher gradients that progressively decrease $\hat{\boldsymbol{\alpha}}_n$. To preserve a high $\boldsymbol{\alpha}_n$ value for the first frame, $\mathbf{M}_n$ is optimized toward 0.

In this way, we can effectively decouple the canonical Gaussians into two groups, as shown in Fig.~\ref{fig:framework}(a), allowing the separation of the highly dynamic foreground (red points) from the background (yellow points) with minimal motion. 

This initial stage not only allows us to learn a better foundational scene representation compared to prior methods like 4D-GS, but the foreground-background decoupling also plays a significant role in subsequent stages, as detailed in the following sections.

\subsection{Stage 2: Large Motion Trajectory Decomposition by  Relay Gaussians}
\label{sec:method:stage2}

Ideally, each foreground Gaussian would follow a large-scale and complex motion trajectory over time, but achieving this directly is challenging. To address this, we replicate multiple copies of the decoupled foreground Gaussians from the first stage, each copy corresponding to a specific temporal segment. In our implementation, consecutive $k$ (\textit{e.g.}, 16) frames are treated as one segment, \textit{i.e.}, the $1^{st}$-$k^{th}$ frames form the first segment, followed by subsequent segments.

Since motion trajectories are continuous over time, these copies, once optimize to the right positions, will break down the large motion trajectory into smaller segments, each representing a portion of the overall motion trajectory. We term them as \textbf{Relay Gaussians} since they serve as explicit relay nodes along the large-scale motion trajectory.

To optimize and densify more Relay Gaussians, we construct pseudo-views by blending $p=3$ uniformly selected frames (\textit{e.g.}, frames 1, 8, and 16 in the first segment) for supervision. Let the three selected frames in the corresponding segment be denoted as $I_{t_1}$, $I_{t_2}$, $I_{t_3}$. The pseudo-view $I_{\text{pseudo}}$ for Relay Gaussians is then constructed as:
\begin{equation}
    I_{\text{pseudo}} = \beta_1 I_{t_1} + \beta_2 I_{t_2} + \beta_3 I_{t_3},
\end{equation}
where $\beta_1 + \beta_2 + \beta_3 = 1$  are blending weights applied to the selected frames, typically chosen based on frame importance or uniform blending. In this work, we use the strightforward uniform blending, \textit{i.e.}, $\beta_1=\beta_2=\beta_3=\frac{1}{3}$, for conciseness. These pseudo-views capture snapshots of the foreground at different time steps, as shown in Fig.~\ref{fig:framework} (b), providing a richer representation for optimizing the Relay Gaussians, ensuring they more accurately capture the motion trajectory within each segment.

By leveraging Relay Gaussians to decompose large-scale motion trajectories into smaller, more manageable segments, we reduce the complexity of handling dynamic motions, which will become evident in the final learning stage. In Sec.~\ref{sec:supp1} of supplementary document, \textit{we analyze this stage as a process of temporal densification}.

\subsection{Stage 3: 4D Spatiotemporal Modeliing and Optimization}
\label{sec:method:stage3}

After the previous two stages, we have established more refined canonical 3D Gaussians. To achieve a complete 4D representation, it is essential to incorporate temporal variation through an motion field. Although our method is not limited to a specific motion field model, in this work, we adopt the HexPlane and MLPs from 4D-GS due to their efficiency and flexibility in spatiotemporal encoding. However, we have made the following improvements to better capture large and complex motions.

\textbf{Foreground-background isolation.} To avoid overfitting to small motions due to all Gaussians sharing MLPs, we propose a divide-and-conquer strategy. For the background Gaussians, we utilize a dedicated set of MLPs that predict the temporal changes in their positions and other attributes. For the foreground Relay Gaussians, another set of MLPs models their time-varying positions and attributes throughout the motion trajectory, as shown in Fig.~\ref{fig:framework} (c).

\textbf{Position deformation scaling.} 
To enhance the ability to capture complex motion patterns, we introduce a learnable scaling factor $\boldsymbol{\gamma} \in \mathbb{R}^{3}$ for each Relay Gaussian. This factor adjusts the predicted position deformations, allowing the model to accommodate larger motion ranges that the MLP alone may not fully capture. This addition ensures that Relay Gaussians can adapt flexibly to intricate motions beyond the standard MLP predictions.
\begin{equation}
    \boldsymbol{\mu} \leftarrow \boldsymbol{\mu} + (1 + e^{\boldsymbol{\gamma}}) \cdot \Delta\boldsymbol{\mu}. 
\end{equation}

By jointly optimizing the canonical Gaussians and our improved motion field, we achieve a comprehensive 4D scene reconstruction that integrates both spatial and temporal dynamics, ultimately yielding a coherent and precise representation of the entire dynamic scene.

%% file: figure/pipeline.tex
\begin{figure*}[t]
  \begin{center}
    \includegraphics[width=1.0\linewidth]{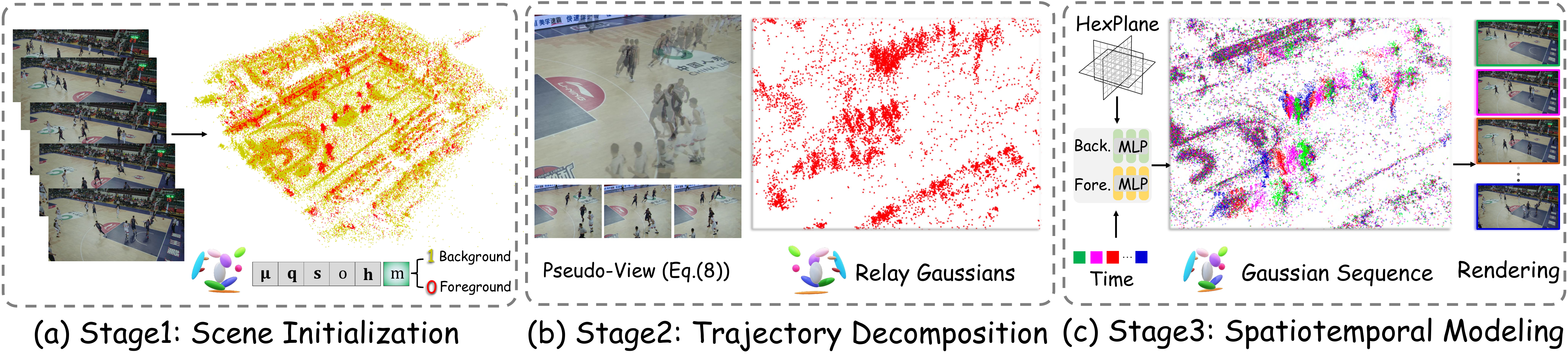}
  \end{center}
  \vspace{-15pt}
  \caption{Framework of the proposed RelayGS. (a) Initialize the scene with all images and separate the relatively static background and dynamic foreground using a learnable mask (visualized as yellow and red). (b) Construct pseudo-GT views through multi-view blending to optimize Relay Gaaussians for decomposing complex trajectories. (c) Based on the HexPlane 4D representation, using different MLPs for foreground and background Gaussians to obtain temporal deformation, and then render through the differentiable pipeline of 3DGS.}
  \label{fig:framework}
\end{figure*}

%% file: sec/5_experiment.tex
\section{Experiment}

\subsection{Experimental Setup}

In this work, we primarily focus on addressing large-scale and complex motion in dynamic scenes. We conduct experiments on the following two representative datasets: \\
\textbf{PanopticSports Dataset} is a subset of the CMU Panoptic Studio dataset~\cite{panoptic_studio}, containing 6 dynamic sports scenes: Juggle, Box, Softball, Tennis, Football and Basketball. Each scene has a resolution of 640$\times$360 and spans 150 frames, captured at 30 FPS. The data was collected using 31 static cameras, of which 27 are used for training and 4 for testing (cameras 0, 10, 15, and 30). \\
\textbf{VRU Basketball Games Dataset}~\cite{vru} contains two real-world basketball game scenes, ``GZ'' and ``DG4''. Each was captured in an indoor basketball court using 34 fixed, synchronized cameras, evenly distributed around the court to cover 360 degrees. The sequences span 10 seconds, with a resolution of 1920$\times$1080 at 25 FPS, resulting in 250 frames per sequence. Of the 34 cameras, 30 are used for training, while 4 (cameras 0, 10, 20, and 30) are reserved for testing.
More details of datasets can be found in the Appendix.

\noindent\textbf{Implementation}.  
Our implementation is based on the open-source 4D-GS~\cite{4dgs} code. In the first stage, 3D Gaussians are initialized using sparse point clouds derived from the initial frame, following 3DGS~\cite{3dgs} and 4D-GS. Each Gaussian is assigned a learnable mask attribute initialized to 2, resulting in values near 1 after sigmoid activation. Optimization runs for 3,000 steps with periodic densification.
In the second stage, we set $k = 16$ and train for 14,000 steps.
In the third stage, we initialize HexPlane and MLPs following 4D-GS, with the difference that we configure two separate MLP sets: one for background Gaussians and the other for Relay Gaussians. Each set of MLPs is responsible for predicting the temporal changes in the four Gaussian attributes—position, scaling, rotation, and opacity. We exclude the spherical harmonics MLP, as it increases model size and reduces rendering speed without significant performance gains. Additionally, the $\boldsymbol{\gamma}$ is initialized to 0. This last stage is trained for 20,000 steps. For the PanopticSports dataset, multi-view color inconsistencies are present, so we apply a learnable channel-wise affine color tune for each camera, following Dynamic3DGS~\cite{dynamic3dgs}. For VRU scenes, we use $2\times$ downsampled views during the first two stages to reduce computational time. All experiments were conducted on an NVIDIA RTX 4090 GPU with batch size 4. The learning rate and densification settings are consistent across all three stages, more details can be found in the Appendix.

\input{table/benchmark_vru}

\input{table/benchmark_panoptic}

\input{figure/quality}

\subsection{Experimental Results}

\paragraph{Quantitative Comparison.}
We compare our RelayGS with several state-of-the-art methods, including 4D-GS~\cite{4dgs}, Dynamic3DGS~\cite{dynamic3dgs}, ST-GS~\cite{st-gs}, E-D3DGS~\cite{e-d3dgs}, and D-MiSo~\cite{d-miso}. The results are shown in Tab.~\ref{tab:benchmark_vru} and Tab.~\ref{tab:benchmark_panoptic}. 
\textbf{(1) Quality}: Our RelayGS method consistently outperforms competitors in terms of reconstruction quality (\textit{i.e.}, PSNR) on both datasets. Specifically, on the six scenes of the PanopticSports dataset (see Tab.~\ref{tab:benchmark_panoptic}), RelayGS achieves PSNR improvements of 0.27 dB, 0.53 dB, 1.6 dB, 1.19 dB, 1.24 dB, and 1.28 dB, respectively, averaging a gain of 1.02 dB over the previous best methods. Compared to the baseline method 4D-GS, we achieve an average performance gain of 2.47 dB. On the more challenging VRU Basketball Games dataset (see Tab.~\ref{tab:benchmark_vru}), RelayGS outperforms the previous best method ST-GS and the baseline method 4D-GS by an average of 0.45 dB and 2 dB, respectively. It is worth \textit{\textbf{noting}} that, although the PSNR difference compared with ST-GS appears small, the static floor occupies approximately 70\% of the pixels in these VRU view images, meaning the quality improvement is more significant in the dynamic foreground regions. Additionally, ST-GS is heavily dependent on initialization, as it extracts sparse point clouds for each frame and then merges them as the initial scene. Since point clouds for each frame cannot be obtained in the PanopticSports dataset, ST-GS is not applicable.
\textbf{(2) Efficiency}: While our method learns corresponding foreground content for each segment via Relay Gaussians, RelayGS strikes a good balance between reconstruction quality and efficiency factors such as storage, training time, and rendering speed compared to competitors, some of which achieve high storage efficiency but fall short in reconstruction quality. In contrast, our method demonstrates a clear advantage in storage efficiency, particularly on the PanopticSports dataset. Compared to the baseline 4D-GS, RelayGS introduces an additional stage with Relay Gaussians, which increases the training time and slightly reduces the rendering speed in some tend. However, RelayGS still maintains a clear advantage in training time compared to other methods. While achieving high-quality reconstruction, we can also ensure a real-time rendering speed of around 70 fps on RTX 4090 GPU.

\paragraph{Qualitative Analysis}
Fig.~\ref{fig:quality_vru_gz} and Fig.~\ref{fig:quality_panoptic_football} show frames from two representative scenes with heavily featured foreground dynamic content. As seen, our RelayGS reconstructs the humans with greater clarity and completeness. This improvement is primarily due to the fact that, compared to our baseline, 4D-GS, our stage I not only learns the background Gaussians but also captures the foreground Gaussians. In our stage II, we further refine the foreground Gaussians by learning additional Gaussians that cover more of the motion trajectories, known as Relay Gaussians. ST-GS, although using point clouds from all 250 frames, obtains a denser sampling of motion trajectories. However, due to its simpler approach to modeling motion changes, it struggles to accurately capture the foreground with complex motions. This issue is more evident in the rendered videos, where ST-GS shows inconsistencies in the motion of the Gaussians associated with the same object, leading to flickering in the foreground. In contrast, our method, leveraging HexPlane encoding following 4D-GS, models temporally and spatially consistent motion, resulting in smoother and more coherent reconstructions.
Additionally, both 4D-GS and E-D3DGS struggle to handle the large-scale motion of the ball in these scenes. In comparison, our method performs significantly better, although challenges remain. The relatively small and isolated ball with mostly empty space around it makes it difficult to track. Our second stage mitigates this issue to some extent by introducing Relay Gaussians, but it remains a challenging aspect due to the sparse Gaussians learned in the first stage.
In summary, RelayGS not only achieves SOTA performance on quantitative metrics but also demonstrates superior spatiotemporal modeling capabilities, particularly on foreground dynamic content. \textit{We encourage readers to view the supplementary rendered videos for a more comprehensive understanding of reconstruction results.}

\vspace{-10pt}

\paragraph{3D Gaussian visualization.}
We visualize the canonical Gaussians learned at different stages, with the results shown in Fig.~\ref{fig:pts_vis}. As observed in Fig.~\ref{fig:pts_vis} (b), in the baseline method 4D-GS, the canonical Gaussians learned in the first stage primarily represent the background, with very few Gaussians capturing the foreground. In contrast, in our method, the base Gaussians learned in the first stage include both background and foreground Gaussians, which can be distinguished by a binary mask, visualized in different colors in Fig.~\ref{fig:pts_vis} (c). Furthermore, through the learning process in the second stage, our method is able to capture additional Relay Gaussians (red points in Fig.~\ref{fig:pts_vis} (d)) along the motion trajectories of the foreground, significantly improving the representation of dynamic content.

\subsection{Ablation Study}
\label{sec:exp:ablation}

In Tab.~\ref{tab:ablation_components}, we present ablation studies on several key components of our method. 
The case \#2 represents the configuration where no foreground Gaussians copies is applied, and only a single global set of foreground Gaussians is used. This results in a significant performance drop, as it cannot effectively handle large-scale motion. 
In case \#3, we remove the second stage of our method, directly replicating a set of foreground Gaussians for each segment and learning them jointly with the implicit motion field. This also leads to a notable performance decrease, especially in the more complex GZ scene. 
In case \#4, we demonstrate the significance of multi-view synthesis pseudo-views, which enable the acquisition of richer Relay Gaussians representing trajectories. In cases \#5 and \#6, we conduct ablation studies on the setting of different MLPs for foreground-background isolation and the scaling factor $\boldsymbol{\gamma}$ in the third stage, respectively. These results highlight the importance of our improvements for 4D spatiotemporal modeling.

In Tab.~\ref{tab:ablation_segment}, we perform an ablation study on the length of each segment, \textit{i.e.}, the number of frames included in each segment. As the segment length increases and the number of segments decreases, the motion trajectory within each segment becomes larger, leading to a gradual decline in performance. However, choosing the $k$ value too small will increase the training cost and not result in a significant performance improvement. Based on experience, we set $k$$=$$16$ as the default selection.

\input{table/ablation}

%% file: table/benchmark_vru.tex
\begin{table*}
\centering  
\caption{Quantitative results on the VRU Basketball Games dataset.  Our RelayGS and other methods only use the point clouds derived from the initial frame. ``ST-GS$^{16}$'' uses point clouds of uniformly selected 16 frames, while ``ST-GS$^{250}$'' utilizes point clouds of all 250 frames, the default setting for their method. Notably, ``ST-GS'' fails to perform effectively when restricted to the same point clouds as ours.}  
\label{tab:benchmark_vru} 
\vspace{-5pt}
\setlength{\tabcolsep}{10.5pt}
\renewcommand{\arraystretch}{1.0}
\begin{tabular}{c cccc cccc}  
    \toprule  
    \multirow{3.5}{*}{Method} &  
    \multicolumn{4}{c}{GZ} & \multicolumn{4}{c}{DG4} \\
    \cmidrule(r){2-5} \cmidrule{6-9} 
    & PSNR  & Storage & Train & Render & PSNR & Storage & Train & Render \\
    & (dB $\uparrow$) & (MB $\downarrow$) & (mins $\downarrow$) & (fps $\uparrow$) & (dB $\uparrow$) & (MB $\downarrow$) & (mins $\downarrow$) & (fps $\uparrow$) \\
    \midrule  
    ST-GS$^{16}$~\cite{st-gs} & \cellthird 26.49 & 35 & 64 & 264 & \cellthird 25.79 & 40 & 64 & 236 \\ 
    \textcolor{gray}{ST-GS$^{250}$}~\cite{st-gs} & \cellsecond \textcolor{gray}{27.32} & \textcolor{gray}{400} & \textcolor{gray}{107} & \textcolor{gray}{143} & \cellsecond \textcolor{gray}{26.79} & \textcolor{gray}{360} & \textcolor{gray}{112} & \textcolor{gray}{134} \\
    \midrule     
    4D-GS~\cite{4dgs} & 25.83 & 42 & 63 & 88 & 25.17 & 45 & 62 & 80 \\ 
    E-D3DGS~\cite{e-d3dgs} & 26.14 & 113 & 224 & 35 & 25.06 & 136 & 301 & 27 \\ 
    \midrule
    \textbf{RelayGS (Ours)} & \cellbest 28.06 & 200 & 105 & 74 & \cellbest 26.94 & 191 & 107 & 69 \\ 
    \bottomrule  
\end{tabular}  
\end{table*}

%% file: table/benchmark_panoptic.tex
\begin{table*}
\centering  
\caption{Quantitative results on the PanopticSports dataset. ``Dynamic3DGS'' and ``D-MiSo'' data are partially taken directly from their original papers or estimated based on the paper and available code. ``Dynamic3DGS'' is a frame-by-frame learning method, whereas our RelayGS and other methods learn from all frames jointly.}
\label{tab:benchmark_panoptic}
\vspace{-5pt}
\setlength{\tabcolsep}{7.3pt}
\renewcommand{\arraystretch}{1.0}
\begin{tabular}{c ccc ccc ccc}  
    \toprule  
    \multirow{3.5}{*}{Method} &  
    \multicolumn{3}{c}{Juggle}&\multicolumn{3}{c}{Boxes}&\multicolumn{3}{c}{Softball} \\
    \cmidrule(r){2-4} \cmidrule{5-7} \cmidrule(r){8-10}
    & PSNR & Storage & Train & PSNR & Storage & Train & PSNR & Storage & Train \\
    & (dB $\uparrow$) & (MB $\downarrow$) & (mins $\downarrow$) & (dB $\uparrow$) & (MB $\downarrow$) & (mins $\downarrow$) & (dB $\uparrow$) & (MB $\downarrow$) & (mins $\downarrow$) \\
    \midrule   
    Dynamic3DGS~\cite{dynamic3dgs} & \cellthird 29.48 & 221 & 107 & \cellsecond 29.46 & 221 & 108  & \cellthird 28.43 & 221 & 116 \\ 
    \midrule
    4D-GS~\cite{4dgs} & 28.19 & 48 & 30 & 27.67 & 47 & 29 & 27.41 & 46 & 29 \\
    E-D3DGS~\cite{e-d3dgs} & 26.54 & 36 & 95 & 26.78 & 33 & 100 & 26.01  & 33 & 80 \\
    D-MiSo~\cite{d-miso} & \cellsecond 29.79 & - & - & \cellthird 29.39 & - & - & \cellsecond 28.60 & - & - \\
    \midrule
    \textbf{RelayGS (Ours)} & \cellbest 30.06 & 31 & 48 & \cellbest 29.99 & 30 & 48 & \cellbest 30.20 & 33 & 48 \\ 
    \toprule  
    & \multicolumn{3}{c}{Tennis} & \multicolumn{3}{c}{Football} & \multicolumn{3}{c}{Basketball} \\ 
    \cmidrule(r){2-4} \cmidrule{5-7} \cmidrule(l){8-10}
     & PSNR & Storage & Train & PSNR & Storage & Train & PSNR & Storage & Train \\
    & (dB $\uparrow$) & (MB $\downarrow$) & (mins $\downarrow$) & (dB $\uparrow$) & (MB $\downarrow$) & (mins $\downarrow$) & (dB $\uparrow$) & (MB $\downarrow$) & (mins $\downarrow$) \\
    \midrule
    Dynamic3DGS~\cite{dynamic3dgs} & \cellthird 28.11 & 221 & 101 & \cellthird 28.49 & 221 & 114  & \cellthird 28.22 & 221 & 113 \\ 
    \midrule
    4D-GS~\cite{4dgs} & 27.49 & 45 & 29 & 26.67 & 54 & 33  & 27.72 & 37 & 24 \\
    E-D3DGS~\cite{e-d3dgs} & 27.41 & 31 & 74 & 25.93 & 33 & 76  & 26.48 & 35 & 87 \\
    D-MiSo~\cite{d-miso} & \cellsecond 29.02 & - & - & \cellsecond 28.99 & - & - & \cellsecond 28.49 & - & - \\
    \midrule
    \textbf{RelayGS (Ours)} & \cellbest 30.21 & 31 & 48 & \cellbest 30.23 & 37 & 48 & \cellbest 29.77 & 51 & 48 \\ 
    \bottomrule  
\end{tabular}  
\end{table*}  

%% file: figure/quality.tex
\begin{figure*}[t]
  \begin{center}
    \includegraphics[width=1.0\linewidth]{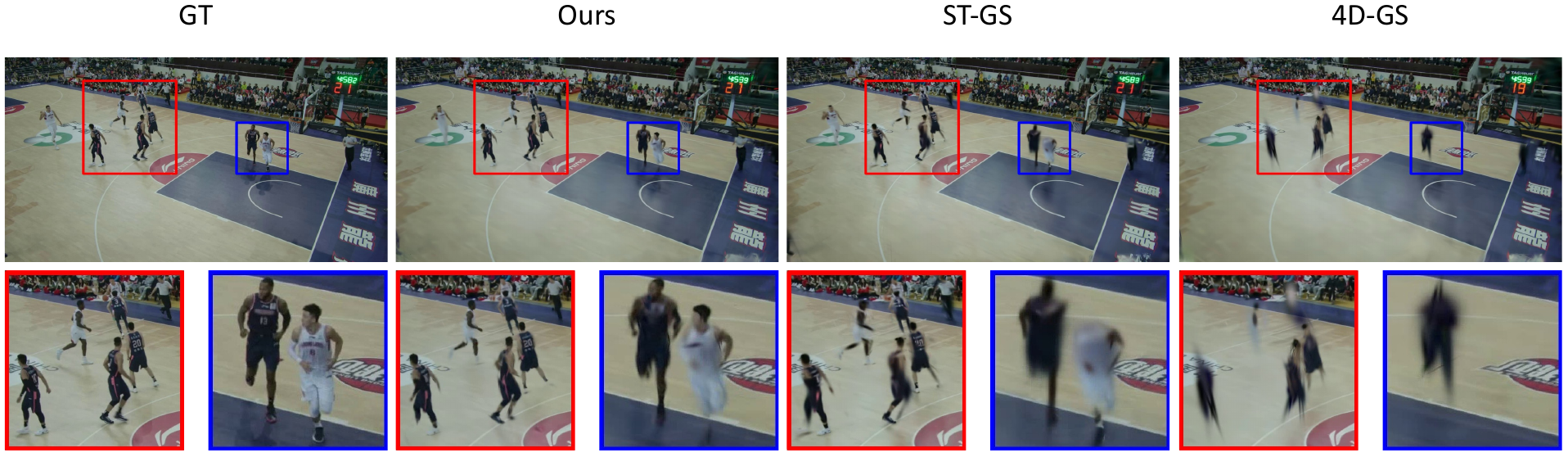}
  \end{center}
  \vspace{-15pt}
  \caption{Qualitative comparisons on GZ scene of VRU Basketball Games dataset. }
  \label{fig:quality_vru_gz}
\end{figure*}

\begin{figure*}[t]
  \begin{center}
    \includegraphics[width=1.0\linewidth]{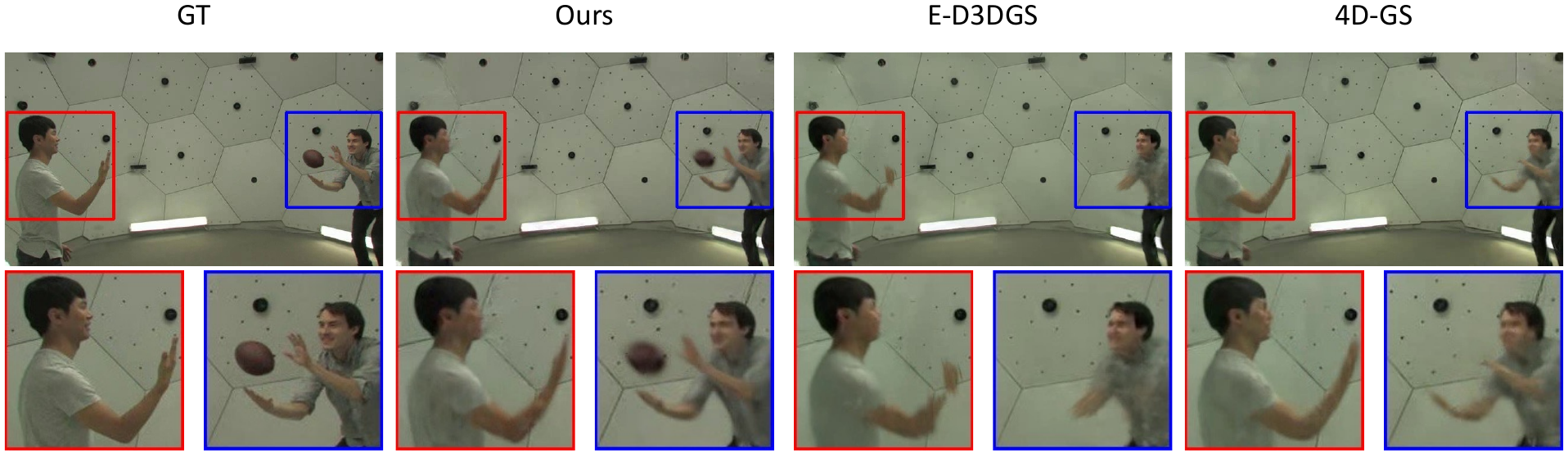}
  \end{center}
  \vspace{-15pt}
  \caption{Qualitative comparisons on Football scene of PanopticSports dataset. }
  \label{fig:quality_panoptic_football}
\end{figure*}

\begin{figure*}[t]
  \begin{center}
    \includegraphics[width=1.0\linewidth]{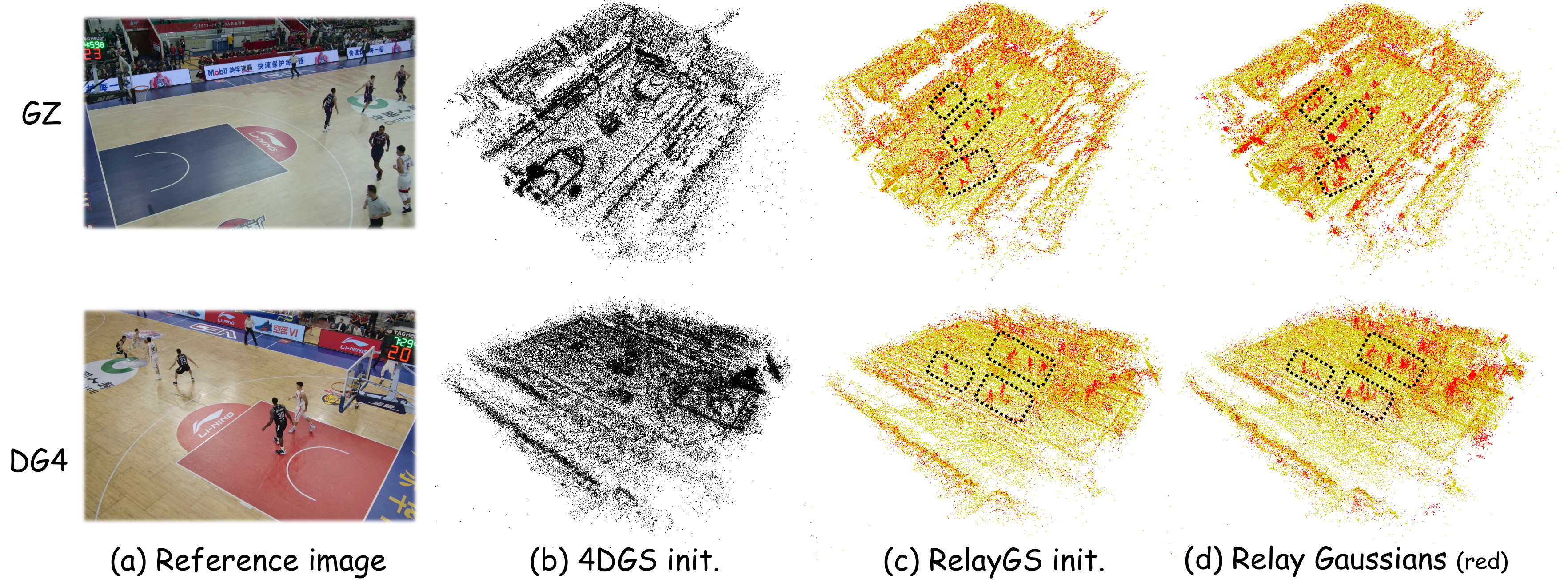}
  \end{center}
  \vspace{-12pt}
  \caption{The visualization of canonical 3D Gaussians. (a) Reference image of the scene. (b) Initialization by 4D-GS, with the foreground Gaussian almost eliminated. (c) Initialization by our method achieves separation of background and foreground, visualized in different colors. (d) Relay Gaussians (red) generated in the second stage realize the decomposition of large-scale complex trajectories.}
  \label{fig:pts_vis}
\end{figure*}

%% file: table/ablation.tex
\begin{table}
\centering  
\caption{Ablation study on key design components. For detailed analysis, please refer to Sec.~\ref{sec:exp:ablation}.}  
\label{tab:ablation_components}  
\setlength{\tabcolsep}{5.4pt}
\renewcommand{\arraystretch}{0.8}
\begin{tabular}{l cc}  
    \toprule  
    Case & GZ & Softball \\
    \midrule
    \#1~~full method & 28.06 & 30.20  \\
    \midrule
    \#2~~\textit{w/o}~ Fg Gaussian Copies & 26.07~\small{\textcolor{red}{$\downarrow$1.99}} & 29.42~\small{\textcolor{red}{$\downarrow$0.78}} \\
    \#3~~\textit{w/o}~ Stage II & 27.27~\small{\textcolor{red}{$\downarrow$0.79}} & 29.93~\small{\textcolor{red}{$\downarrow$0.27}} \\
    \#4~~\textit{w/o}~ Pseudo-Views & 27.80~\small{\textcolor{red}{$\downarrow$0.26}} & 30.00~\small{\textcolor{red}{$\downarrow$0.20}} \\
    \#5~~\textit{w/o}~ Fg-Bg Isolation & 27.80~\small{\textcolor{red}{$\downarrow$0.26}} & 30.07~\small{\textcolor{red}{$\downarrow$0.13}} \\
    \#6~~\textit{w/o}~ Scaling Factor $\boldsymbol{\gamma}$ & 27.87~\small{\textcolor{red}{$\downarrow$0.19}} & 29.73~\small{\textcolor{red}{$\downarrow$0.47}} \\
    \bottomrule  
\end{tabular}  
\end{table}

\begin{table}
\centering  
\caption{Ablation on number of frames per segment.}  
\label{tab:ablation_segment}  
\setlength{\tabcolsep}{6.5pt}
\renewcommand{\arraystretch}{0.8}
\begin{tabular}{cccc cc}  
    \toprule  
    $k$ & 8 & 16 & 32 & 64 & 128  \\
    \midrule     
    PSNR (dB) & 27.90 & 28.06 &  27.82 & 27.56 & 27.10  \\  
    \bottomrule  
\end{tabular}  
\end{table}

%% file: sec/6_conclusion.tex
\section{Conclusion}
This paper introduces RelayGS to tackle the challenges of reconstructing dynamic scenes with large-scale and complex motions. We first learn the basic scene structure and, through a learnable mask, simultaneously capture the shared low-dynamic background and high-dynamic foreground, achieving effective decoupling of foreground and background Gaussians. 
Then, we replicate the foreground Gaussians and train them with pseudo-views constructed by blending frames within corresponding temporal segments. These foreground Gaussians are referred to as Relay Gaussians, which decompose the complex, large-scale motion trajectories into smaller, manageable segments. 
Finally, we jointly optimize a compact motion field and the canonical Gaussians to learn a comprehensive 4D representation of the dynamic scene. Experiments on two real-world datasets demonstrate that RelayGS achieves state-of-the-art reconstruction quality for large-scale motions while maintaining a balance between reconstruction fidelity and storage efficiency, making it practical for real-world dynamic scene applications.

\noindent\textbf{Limitations.} While our method achieves significant performance advantages, it still faces some known challenges. \textit{(1)} Insufficient motion modeling of small but fast-moving objects due to the limited pixel coverage of these objects, insufficient camera view coverage, and sparse canonical surrounding Gaussians. \textit{(2)} Our temporal segmentation and pseudo-view construction strategies are relatively straightforward. In the future, we plan to explore more adaptive temporal segmentation methods that align with the motion complexity of the scene. Moreover, we aim to develop more sophisticated frame selection strategies that better capture motion dynamics, enabling Relay Gaussians to more closely follow the ideal motion trajectory and improve the accuracy of motion representation. \textit{(3)} Our method is tailored for multi-view inputs from stationary cameras and may not be suitable for settings with monocular videos or those involving moving \mbox{cameras}.

%% file: sec/X_suppl.tex
\clearpage
\setcounter{page}{1}
\maketitlesupplementary

This supplementary document provides additional insights and details to support our main paper. In Section~\ref{sec:supp1}, we analyze our Relay Gaussians from the perspective of Gaussian densification. In Section~\ref{sec:supp2}, we emphasize that our method adopts a unified reconstruction framework rather than a segment-based approach, aiming to prevent any potential misunderstandings. Section~\ref{sec:supp3} provides detailed information about the datasets used in our experiments. Section~\ref{sec:supp4} presents more implementation details to facilitate reproducibility. Finally, in Section~\ref{sec:supp5}, we showcase additional experimental results, including qualitative and quantitative comparisons, along with a description of the accompanying videos for better visualization of our method’s performance.

\section{Relay Gaussians from Densification Perspective}
\label{sec:supp1}

\textbf{Spatial Densification.}
In standard 3DGS, regions with insufficient spatial representation are typically addressed by add Gaussians in those areas, an operation we refer to as \textit{spatial densification}. Most prior 4D reconstruction methods~\cite{deformabl3dgs,4dgs} adopt canonical 3D Gaussians combined with a temporal deformation field as the 4D representation framework. Consequently, densification is performed solely in the canonical 3D space, essentially extending the spatial densification strategy of 3DGS into 4D settings, as illustrated in Fig.~\ref{fig:densification}~(a). These methods assume that a single canonical Gaussian corresponds to an entire motion trajectory across time, with the motion field responsible for learning the Gaussian’s position at each time step. However, this assumption is overly idealized and proves challenging in practice. Real-world scenes often involve large-scale, complex motions, and motion fields, typically implict, may struggle to capture these trajectories accurately, leading to significant errors in both spatial and temporal alignment.

\input{figure/densification}

\noindent\textbf{Temporal Densification.}
Analogous to spatial densification in static 3DGS, the idealized motion trajectory of a canonical Gaussian along the time dimension in dynamic scene reconstruction may be underrepresented. An intuitive solution to this issue, as depicted in Fig.~\ref{fig:densification}~(b), is to introduce additional Gaussians along the trajectory and optimize them, progressively achieving a more accurate representation of the intended motion trajectory. This operation is referred to by us as \textit{temporal densification}.

In the second stage of our method, multiple copies of the foreground Gaussians are replicated and optimized to their target positions, which fundamentally constitutes \textit{temporal densification}. This process directly increases the temporal density of the Gaussians along their idealized motion trajectories, ensuring they are sufficiently distributed to capture the complex dynamics of highly active regions. By doing so, it lays the groundwork for accurately representing large-scale, dynamic motions.

However, in dynamic scenes, the non-rigid nature of dynamic objects introduces further complexity. The same object may undergo varying transformations at different time steps, sometimes requiring more Gaussians for accurate representation, sometimes fewer, and occasionally none at all—such as when the object moves out of the scene or is occluded or enveloped by other content. Temporal densification must adapt to these variations, ensuring that the Gaussian representations dynamically align with the scene’s temporal and structural changes for optimal fidelity and efficiency. 

Thus, the second stage of our method is designed as a dedicated process that goes beyond simple temporal densification by leveraging pseudo-views constructed from multiple temporal frames to refine the process. These pseudo-views serve two critical purposes: first, they enable a more intensive temporal densification by providing additional supervisory signals, ensuring that the replicated Gaussians are further aligned with complex and large-scale motion trajectories. Second, they support enhanced spatial densification by guiding the optimization of Gaussians to adapt to non-rigid transformations. This ensures that the learned canonical Gaussians achieve both temporal precision and spatial consistency, providing a robust and unified foundation for accurate and adaptable dynamic scene reconstruction.

\section{Unified \textit{vs.} Segment-Based Reconstruction}
\label{sec:supp2}

Despite the explicit use of temporal segments in the second stage of our method for learning Relay Gaussians, our approach is fundamentally different from segment-based reconstruction methods such as Deformable3DGS~\cite{deformabl3dgs} and ST-GS~\cite{st-gs}, which rely on segment-wise learning for long-term dynamic scenes.

Segment-based methods treat each temporal segment as an independent learning task, reconstructing a motion field and 4D representation for each segment separately. In contrast, our approach leverages temporal segmentation purely as an optimization strategy within a unified framework, where all segments collectively contribute to a single, cohesive 4D representation.

Instead of performing multiple independent reconstructions—e.g., 10 separate processes for a 250-frame sequence divided into 25-frame segments—our method employs a single training pipeline across three fixed stages (3k, 14k, and 20k steps, respectively). This unified process not only ensures temporal coherence across the entire sequence but also significantly reduces reconstruction time and storage requirements compared to segment-based methods.

\section{Dataset Details}
\label{sec:supp3}

\textbf{PanopticSports Dataset.}
The cameras are temporally aligned with accurate intrinsic and extrinsic parameters. Positioned in a roughly hemispherical arrangement around the area of interest in the middle of the capture studio, the cameras provide comprehensive coverage of the scene. The images are undistorted using the provided distortion parameters and resized to 640 × 360. The dataset provides a point cloud generated by 10 available depth cameras for each scene. In our experiments, this point cloud is first downsampled to approximately 35,000 points, which are then used to initialize the Gaussian primitives. Each scene involves one or two moving persons and some moving objects, while the background remains completely static. Additionally, the foreground colors are quite similar to the background, which further increases the reconstruction difficulty due to the reduced contrast between the foreground and background.

\noindent\textbf{VRU Basketball Games Dataset.} 
The camera poses and distortion parameters were estimated using the first frame from all 34 views by COLMAP~\cite{sfm}, and all frames were undistorted accordingly. After undistortion, the resolution slightly increases, and we did not resize the images back to 1920$\times$1080. Following the 4D-GS~\cite{4dgs} method, a point cloud was generated and downsampled to approximately 80,000 points for initializing the Gaussian primitives. Each scene includes multiple basketball players, a basketball, scoreboards, advertisement banners, and thousands of spectators. The basketball players and the basketball exhibit fast and large-scale movements with highly complex motion patterns, including non-rigid deformations. The scoreboards and banners also dynamically change over time, and even the background spectators are not completely static, as some exhibit subtle movements. Additionally, the physical scale of the scene is significantly larger than previously available dynamic scene datasets, making it highly challenging to reconstruct.

\section{More Implementation Details}
\label{sec:supp4}

Our method employs slightly different settings for learning rates and densification thresholds between the foreground and background Gaussians. The background learning rates are similar to those used in previous methods, with the initial learning rate for position set to 2e-4 and the minimum learning rate to 1e-5. For the foreground Gaussians, the initial learning rate for position is set to 1e-3. The gradient threshold for densification is 1e-4, which is half of the threshold used for the background. Additionally, the scaling threshold for densification is set to 1e-3 for the foreground, which is 0.1 times that of the background. These settings encourage the foreground Gaussians to be smaller and split faster than the background Gaussians.
More detailed experimental settings will be released in our future open-source code to better support reproducible research.

\input{figure/quality_appendix}

\input{table/benchmark_vru_half_res}

\section{Additional Experimental Results}
\label{sec:supp5}

The goal of the first two stages of our method is to learn a more robust base Gaussian representation, simplifying complex motion patterns in the scene and preparing for full learning in the final stage. Using low-resolution views during these stages produces comparable results while significantly reducing training time. Additionally, we observed that our method performs more effectively at low resolutions, resulting in a larger performance gap compared to counterpart methods. The results are presented in Tab.~\ref{tab:benchmark_vru_half_res}, further reinforcing the superiority of our approach in motion learning.

We present the quality comparison on other scenes from the two datasets in Figures~\ref{fig:quality_vru_dg4} to ~\ref{fig:quality_panoptic_basketball}. To highlight the differences between our method and other methods, we have marked specific foreground regions with red and blue boxes and magnified them for closer inspection. As shown in the magnified views, our method reconstructs the foreground more completely and produces higher-quality details, demonstrating superior performance in preserving fine-grained structures. These qualitative results clearly demonstrate that our method consistently achieves significantly better visual quality compared to competitive counterparts across different scenes from both datasets, highlighting the generalization ability and robustness of our RelayGS method.

In Fig.~\ref{fig:relay_gaussians_panoptic}, we provide additional visualization results of Relay Gaussians on the PanopticSports dataset, showcasing how our method learns Relay Gaussians for large-scale dynamic content.

In the zip file of this Supplementary Material, which includes this document, there are 3 videos (also accessible \href{https://github.com/gqk/RelayGS}{online}), all composed of 4 test views with 10 seconds per view, resulting in a total duration of 40 seconds:
\begin{itemize}
    \item \texttt{VRU\_GZ\_GT.mp4}: The ground truth video.
    \item \texttt{VRU\_GZ\_RelayGS\_PSNR-28.06.mp4}: The video rendered by our RelayGS method. 
    \item \texttt{VRU\_GZ\_ST-GS\_PSNR-27.32.mp4}: The video rendered by the prior SOTA ST-GS~\cite{st-gs} method initialized using the sparse point clouds of all 250 frames.
\end{itemize}
These videos allow a direct comparison of reconstruction quality and motion coherence. As shown, our RelayGS method demonstrates superior performance in both aspects compared to the competitive ST-GS method.

%% file: figure/densification.tex
\begin{figure}[t]
  \begin{center}
    \includegraphics[width=1.0\linewidth]{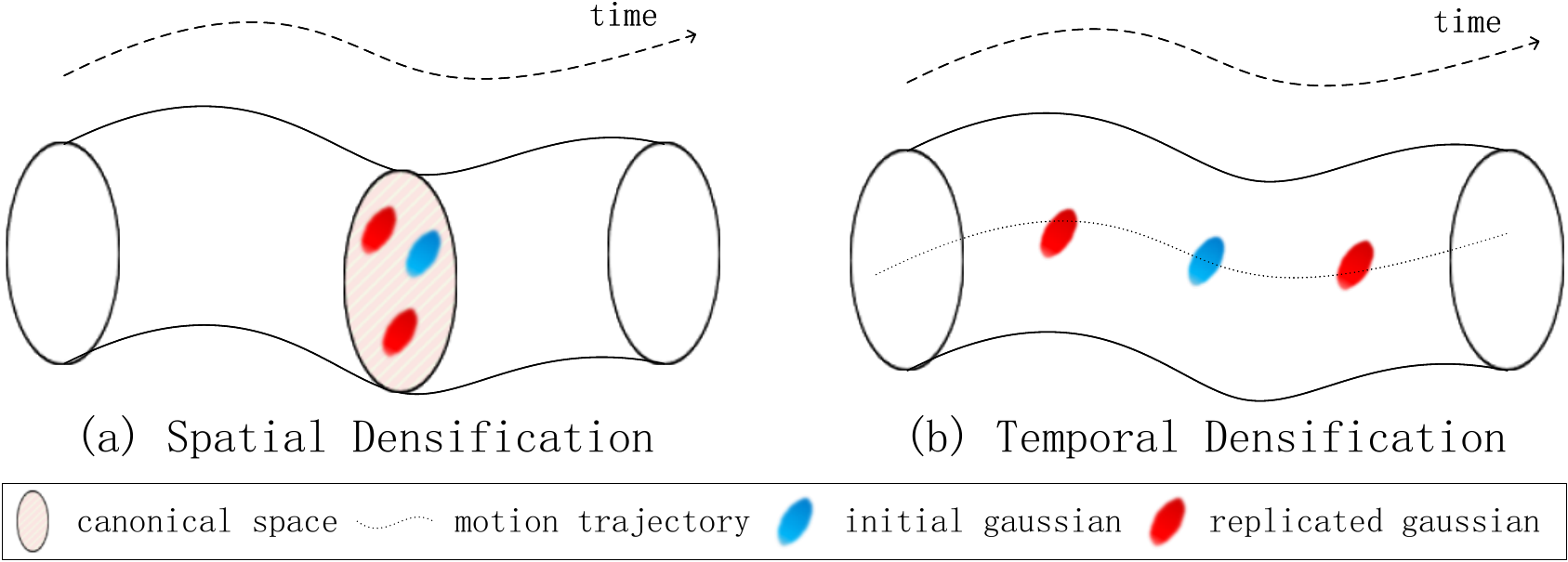}
  \end{center}
  \vspace{-3mm}
  \caption{\textbf{Illustrative depiction of two types of densification.} In 3DGS for static scene reconstruction, \textit{spatial densification} is employed to better fit 3D structures. Prior 4D methods, as shown in (a), perform densification within a canonical 3D space, relying on deformation fields to model motion trajectories, but often fail to sufficiently represent these trajectories. As shown in (b), explicitly densifying along the motion trajectory by adding new Gaussians enables a more accurate representation of dynamic motion. Our method introduces Relay Gaussians, fundamentally rooted in the intrinsic combination of spatial and temporal densification, enabling enhanced 4D reconstruction.}
  \label{fig:densification}
\end{figure}

%% file: figure/quality_appendix.tex
\begin{figure*}[ht]
  \begin{center}
    \includegraphics[width=1.0\linewidth]{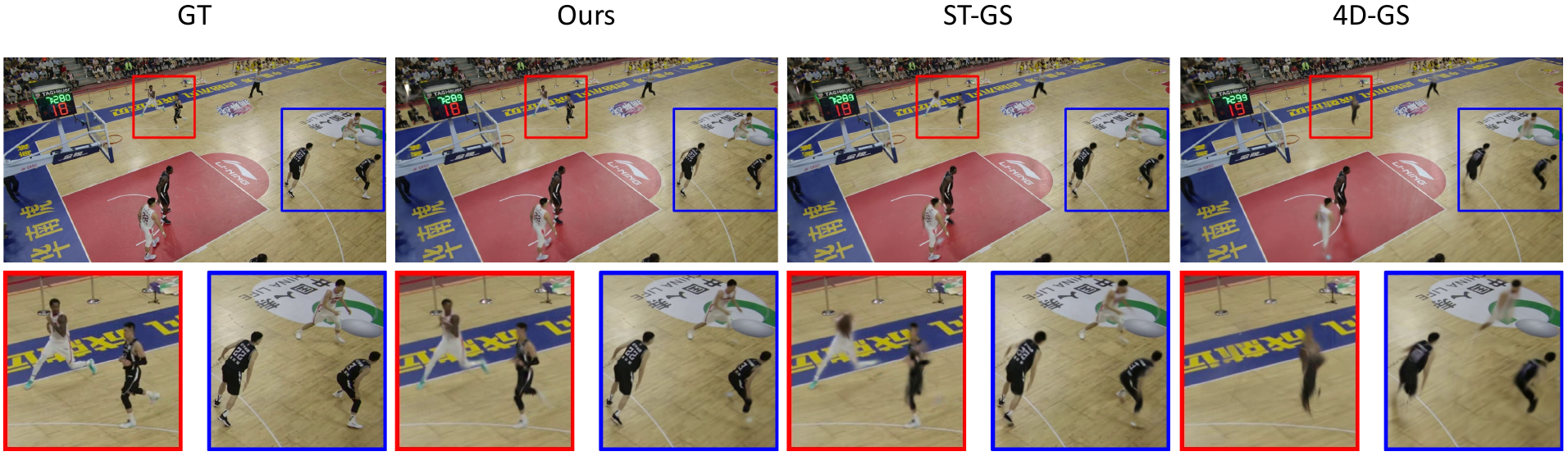}
  \end{center}
  \vspace{-3mm}
  \caption{Qualitative comparisons on DG4 scene of VRU Basketball Games dataset. }
  \label{fig:quality_vru_dg4}
\end{figure*}

\begin{figure*}[ht]
  \begin{center}
    \includegraphics[width=1.0\linewidth]{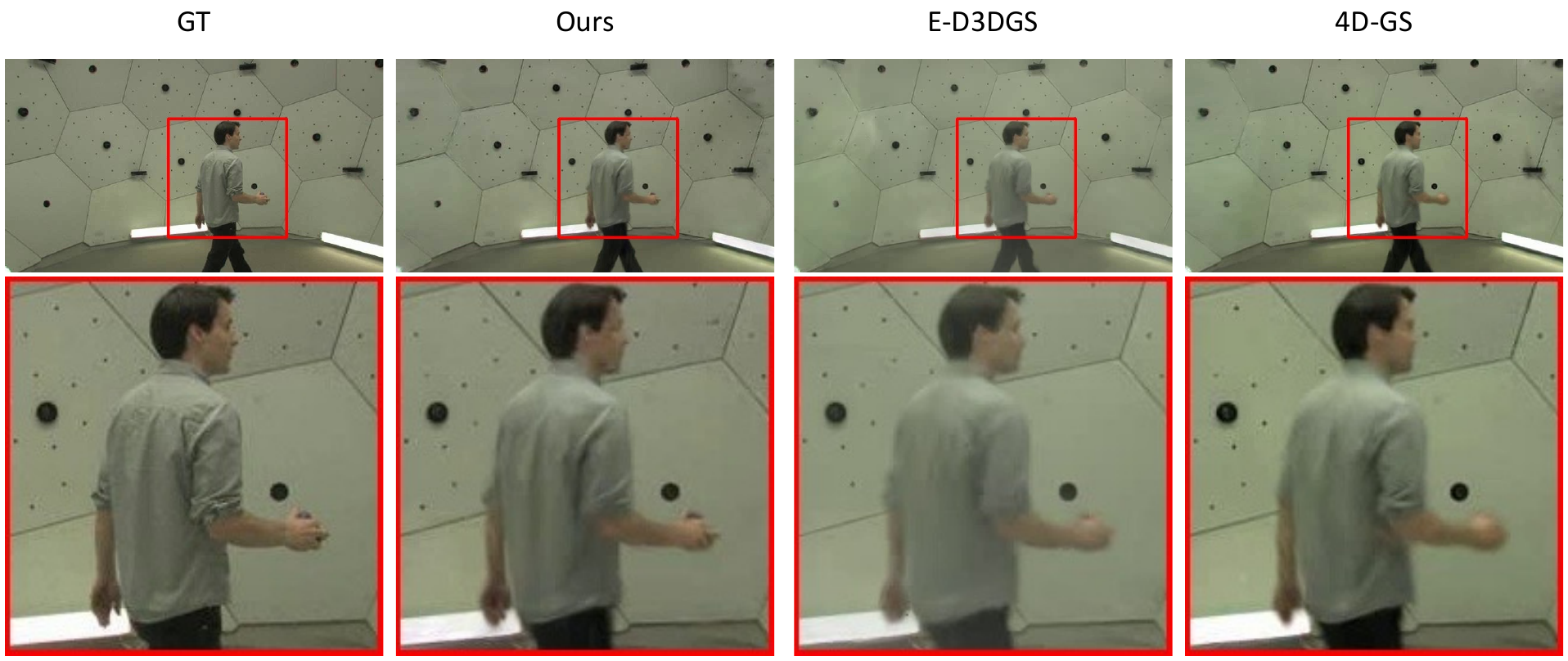}
  \end{center}
  \vspace{-3mm}
  \caption{Qualitative comparisons on Juggle scene of PanopticSports dataset. }
  \label{fig:quality_panoptic_juggle}
\end{figure*}

\begin{figure*}[ht]
  \begin{center}
    \includegraphics[width=1.0\linewidth]{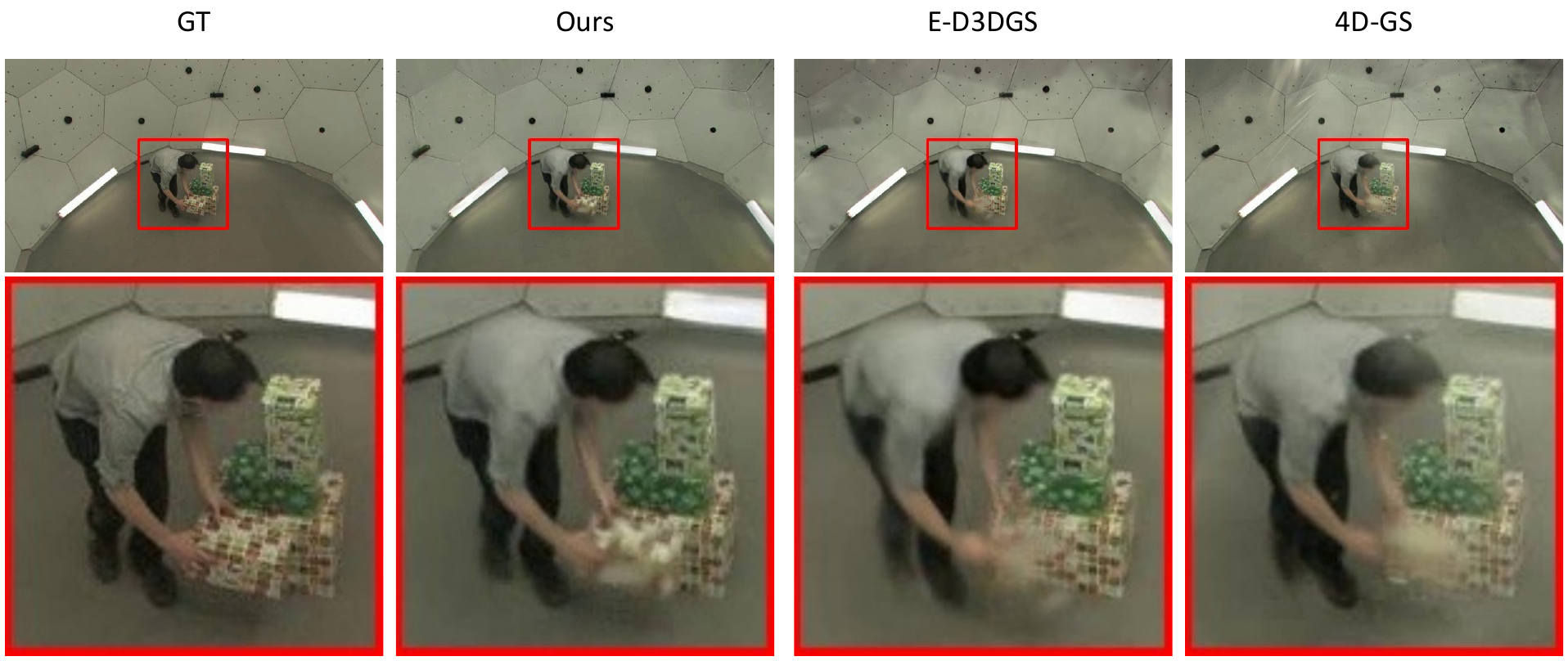}
  \end{center}
  \vspace{-3mm}
  \caption{Qualitative comparisons on Boxes scene of PanopticSports dataset. }
  \label{fig:quality_panoptic_boxes}
\end{figure*}

\begin{figure*}[ht]
  \begin{center}
    \includegraphics[width=1.0\linewidth]{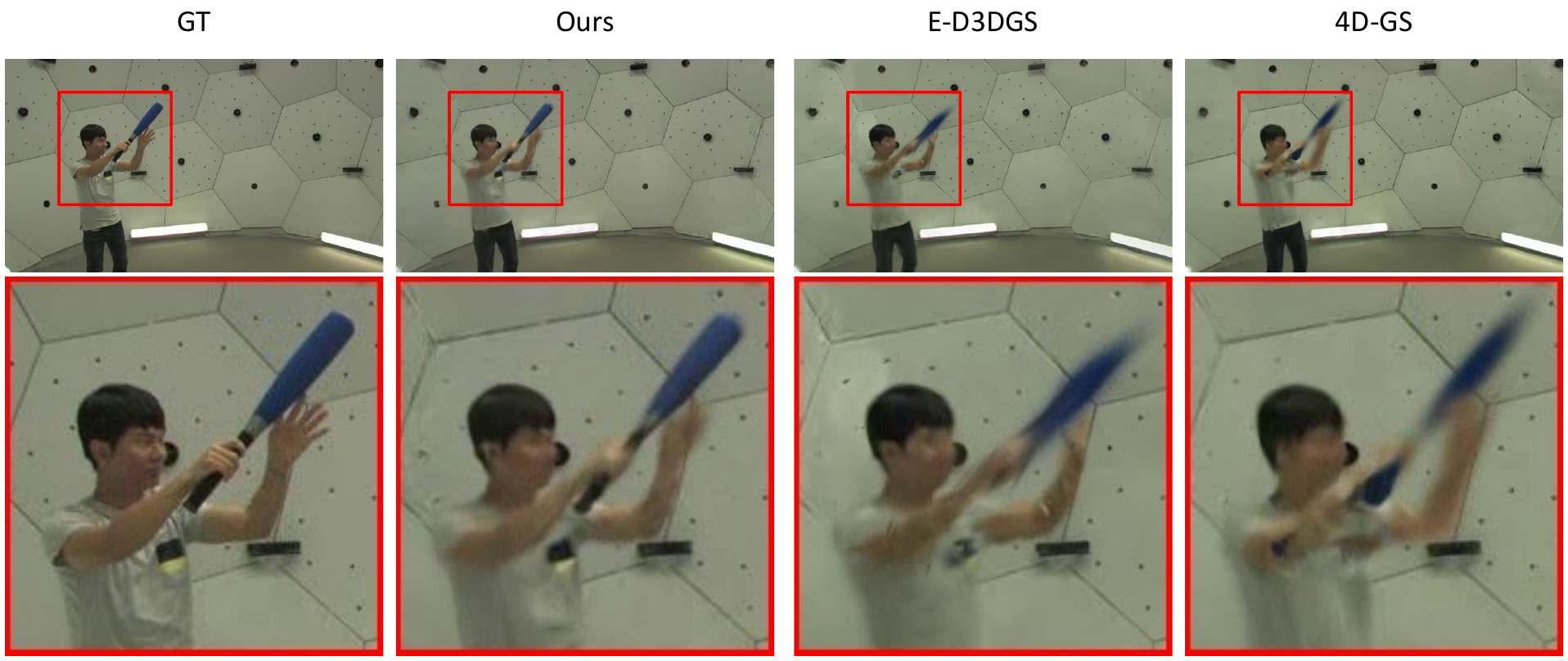}
  \end{center}
  \vspace{-3mm}
  \caption{Qualitative comparisons on Softball scene of PanopticSports dataset. }
  \label{fig:quality_panoptic_softball}
\end{figure*}

\begin{figure*}[ht]
  \begin{center}
    \includegraphics[width=1.0\linewidth]{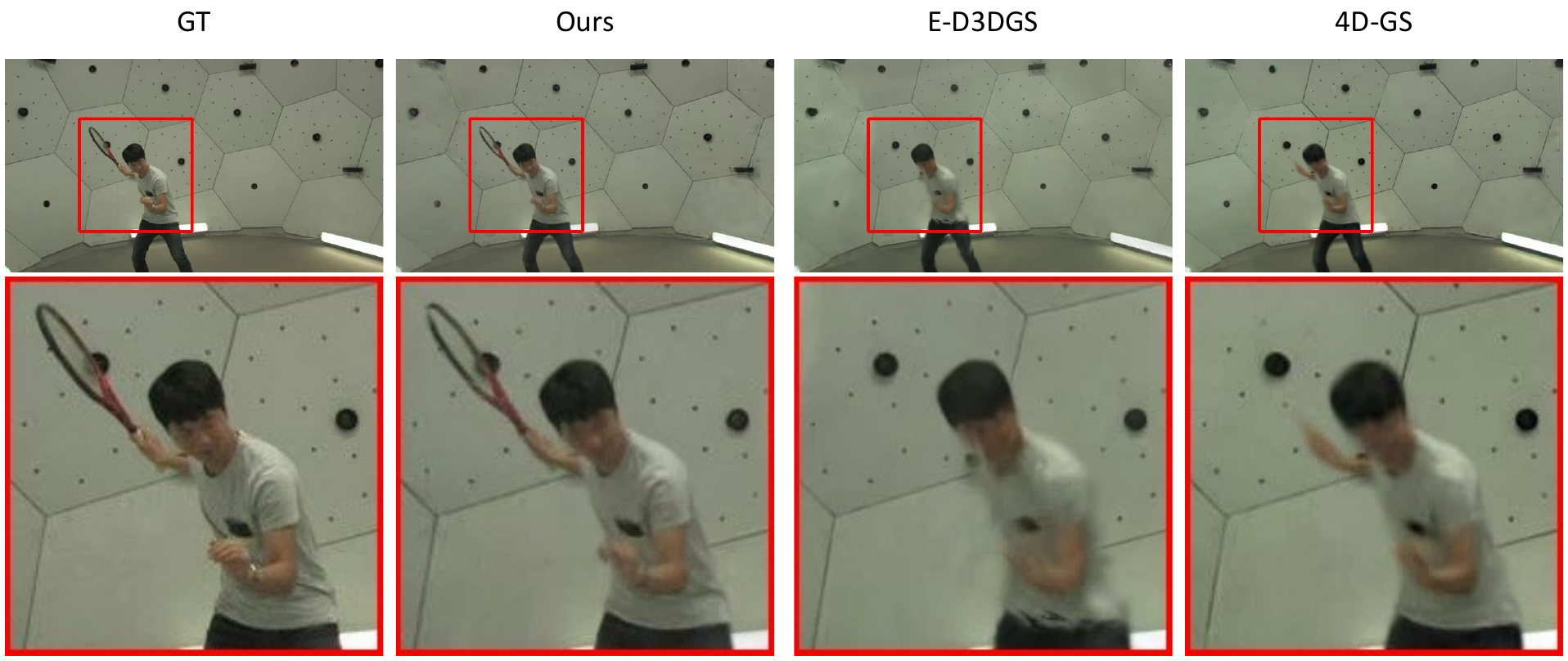}
  \end{center}
  \vspace{-3mm}
  \caption{Qualitative comparisons on Tennis scene of PanopticSports dataset.}
  \label{fig:quality_panoptic_tennis}
\end{figure*}

\begin{figure*}[ht]
  \begin{center}
    \includegraphics[width=1.0\linewidth]{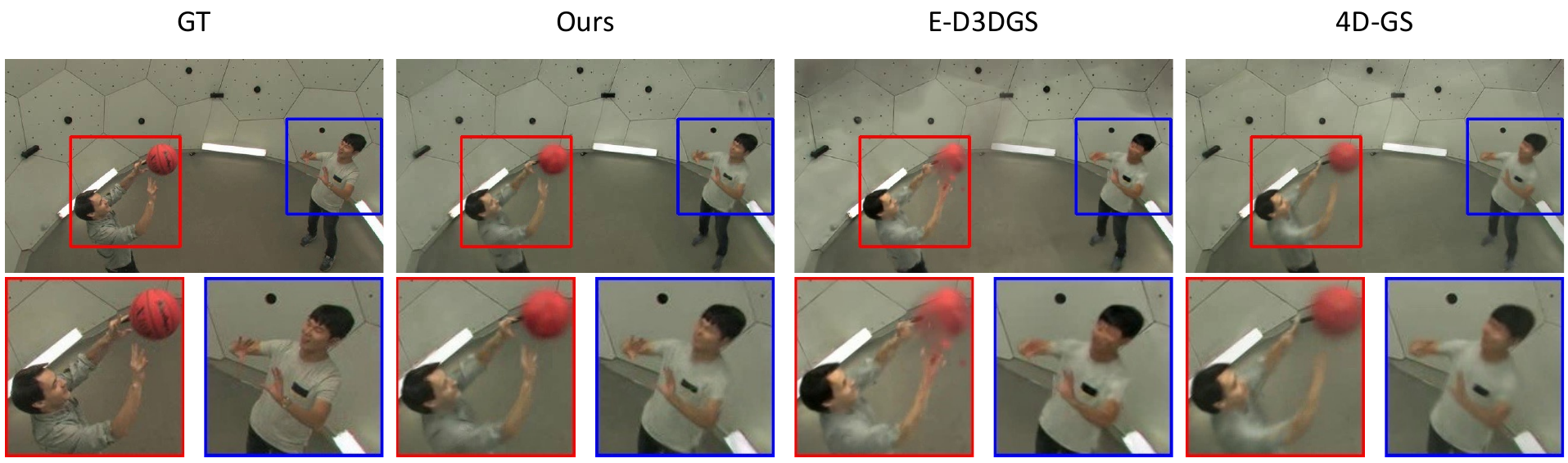}
  \end{center}
  \vspace{-3mm}
  \caption{Qualitative comparisons on Basketball scene of PanopticSports dataset. }
  \label{fig:quality_panoptic_basketball}
\end{figure*}

\begin{figure*}[ht]
  \begin{center}
    \includegraphics[width=1.0\linewidth]{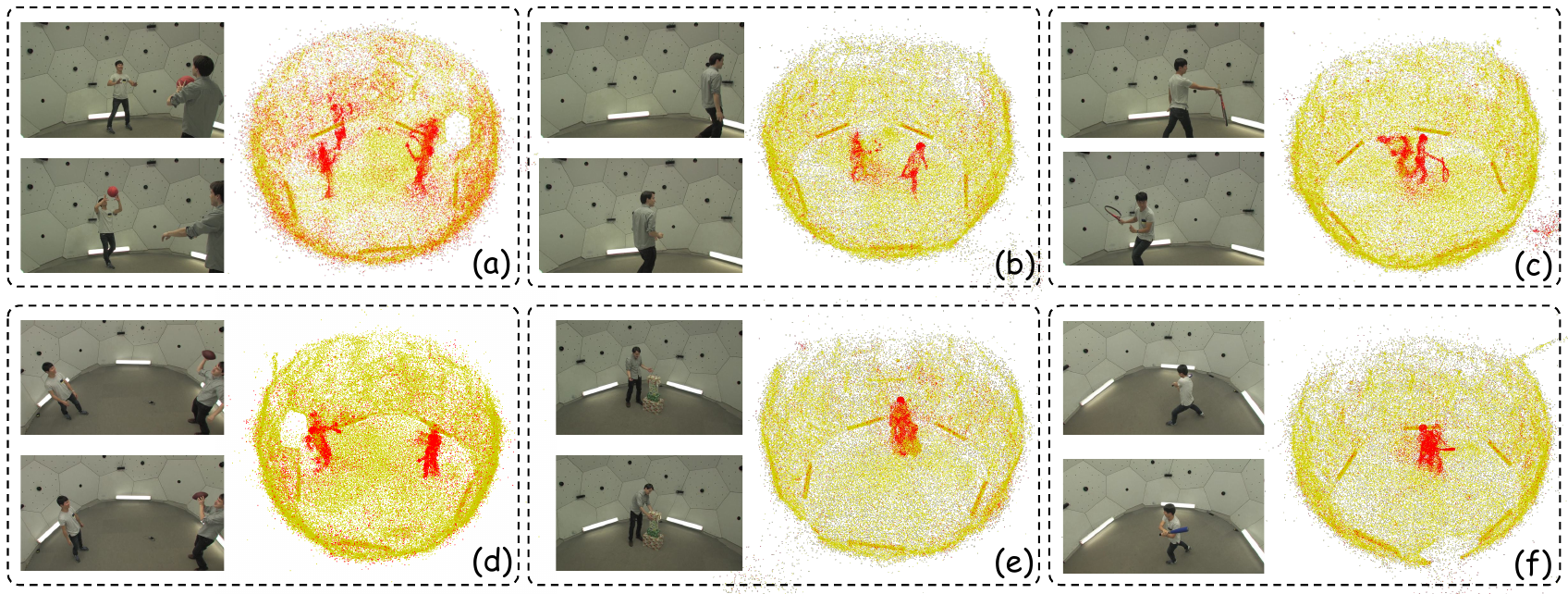}
  \end{center}
  \vspace{-3mm}
  \caption{Visualizations of the second-stage dynamic foreground Relay Gaussians (red points) in 6 scenes of the PanopticSports dataset. (a)-(c) show people in the foreground with larger motion amplitudes, generating more dispersed trajectories. (d)-(f) show people in the foreground with smaller motion amplitudes, generating more concentrated trajectories.}
  \label{fig:relay_gaussians_panoptic}
\end{figure*}

%% file: table/benchmark_vru_half_res.tex
\begin{table}
\centering  
\caption{Quantitative results on the VRU Basketball Games dataset at half resolution. ``ST-GS'' utilizes point clouds of all 250 frames, the default setting for their method.}  
\label{tab:benchmark_vru_half_res} 
\setlength{\tabcolsep}{20pt}
\begin{tabular}{c cccc}  
    \toprule  
    \multirow{2}{*}{Method} & \multicolumn{2}{c}{PSNR  (dB $\uparrow$)} \\
    \cmidrule(r){2-3}
    & GZ & DG4 \\
    \midrule     
    ST-GS~\cite{st-gs} & \cellsecond 27.61  & \cellsecond 26.87 \\  
    E-D3DGS~\cite{e-d3dgs} & 26.33  & 25.39  \\ 
    \midrule
    \textbf{RelayGS (Ours)} & \cellbest 28.97  & \cellbest  27.50 \\ 
    \bottomrule
\end{tabular}  
\end{table}